\documentclass[10pt,twocolumn,letterpaper]{article}
\usepackage[accsupp]{axessibility} 
\usepackage[pagenumbers]{wacv} 

\usepackage{graphicx}
\usepackage{amsmath}
\usepackage{amssymb}
\usepackage{booktabs}
\usepackage{array}
\usepackage{multirow}

\usepackage[pagebackref,breaklinks,colorlinks]{hyperref}

\usepackage[capitalize]{cleveref}
\crefname{section}{Sec.}{Secs.}
\Crefname{section}{Section}{Sections}
\Crefname{table}{Table}{Tables}
\crefname{table}{Tab.}{Tabs.}

\def\wacvPaperID{1916} 
\def\confName{WACV}
\def\confYear{2025}

\begin{document}

\title{MLLM-LLaVA-FL: Multimodal Large Language Model Assisted Federated Learning}

\author{First Author\\
Institution1\\
Institution1 address\\
{\tt\small firstauthor@i1.org}
\and
Second Author\\
Institution2\\
First line of institution2 address\\
{\tt\small secondauthor@i2.org}
}

\author{
\textbf{Jianyi Zhang}\textsuperscript{1},
\textbf{Hao Yang}\textsuperscript{2},
\textbf{Ang Li}\textsuperscript{3},
\textbf{Xin Guo}\textsuperscript{4},
\textbf{Pu Wang}\textsuperscript{2},
\textbf{Haiming Wang}\textsuperscript{4},\\
\textbf{Yiran Chen}\textsuperscript{1},
\textbf{Hai Li}\textsuperscript{1},
\\
  \textsuperscript{1} Duke University,
  \textsuperscript{2} Johns Hopkins University, \\
  \textsuperscript{3} University of Maryland College Park
  \textsuperscript{4} Lenovo Research
\\
}

\maketitle

\begin{abstract}
   Previous studies on federated learning (FL) often encounter performance degradation due to data heterogeneity among different clients. In light of the recent advances in multimodal large language models (MLLMs), such as GPT-4v and LLaVA, which demonstrate their exceptional proficiency in multimodal tasks, such as image captioning and multimodal question answering. We introduce a novel federated learning framework, named Multimodal Large Language Model Assisted Federated Learning (MLLM-LLaVA-FL), which employs powerful MLLMs at the server end to address the heterogeneous and long-tailed challenges. Owing to the advanced cross-modality representation capabilities and the extensive open-vocabulary prior knowledge of MLLMs, our framework is adept at harnessing the extensive, yet previously underexploited, open-source data accessible from websites and powerful server-side computational resources. Hence, the MLLM-LLaVA-FL not only enhances the performance but also avoids increasing the risk of privacy leakage and the computational burden on local devices, distinguishing it from prior methodologies. Our framework has three key stages. Initially, we conduct global visual-text pretraining of the model. This pretraining is facilitated by utilizing the extensive open-source data available online, with the assistance of MLLMs. Subsequently, the pretrained model is distributed among various clients for local training. Finally, once the locally trained models are transmitted back to the server, a global alignment is carried out under the supervision of MLLMs to further enhance the performance. Experimental evaluations on established benchmarks, show that our framework delivers promising performance in the typical scenarios with data heterogeneity and long-tail distribution across different clients in FL. 
\end{abstract}

\section{Introduction}
\label{sec:intro}
\begin{figure*}[ht]
\vspace{-1cm}
    \centering
    \includegraphics[width=0.8\linewidth]{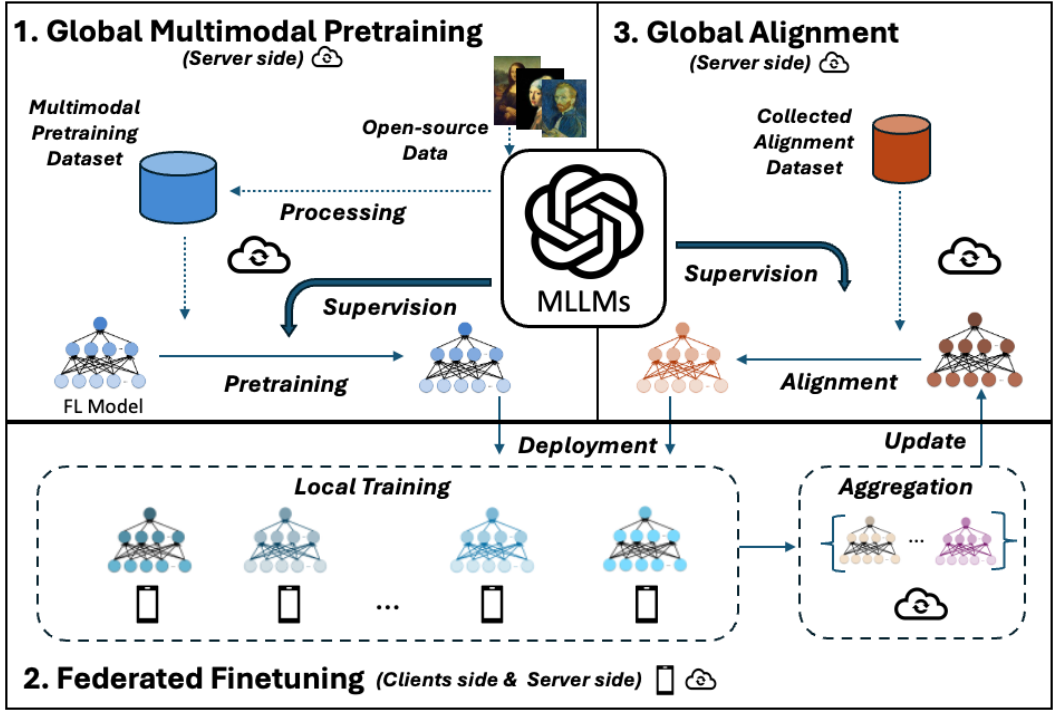}
    \caption{The workflow of MLLM-LLaVA-FL. The MLLM are utilized in the first stage Global Multimodal Pretraining and the third stage Global Alignment on the server side, to avoid extra computational load on devices.}
    \label{fig:framework1}
\end{figure*}

Data heterogeneity represents a significant challenge in federated learning\cite{mcmahan2016communication, kairouz2021advances,sahu2018federated,zhang2024towards,hao2021towards,zhang2022next}. It largely stems from the fact that the data across participating clients are distributed independently, with each client having a different sample distribution. Due to the diversity in clients' datasets, these datasets often exhibit a long-tailed distribution, leading to client models that are biased toward the more common classes \cite{sahu2018federated,wang2020addressing}. This discrepancy often results in a drop in model accuracy. Although several approaches have been proposed \cite{li2022effectiveness,huang2021behavior,huang2023rethinking,feng2023towards,shi2023clip,flop_qian2021,jia2025unlocking,du2022rethinking}, the majority fail to strike an optimal balance between performance and mitigating two critical issues: 1. avoiding concerns of privacy leakage, and 2. preventing the imposition of extra computational loads on local edge devices. For instance, some contemporary methodologies necessitate the transmission of both gradients and parameters from local models to the server, which introduces substantial privacy risks. This is because attackers could potentially reverse-engineer the transmitted data to reconstruct client-specific images, as highlighted in various studies \cite{geiping2020inverting,zhu2019deep,haim2022reconstructing}. Alternatively, other approaches require the deployment of sizable models on local devices, which increases the memory and computational demands.

In light of the current popularity and exceptional proficiency of multimodal large language models (MLLMs) in tasks involving multimodalities, such as image captioning and multimodal question answering, we introduce a three-stage framework, named Multimodal Large Language Model
sisted Federated Learning (MLLM-LLaVA-FL), which utilizes multimodal large language models (MLLMs) to the FL performance on heterogeneous and long-tailed data. 
The adaptation of MLLMs in FL is supported by two main considerations. Firstly, beyond the heterogenous and long-tailed distribution of client datasets, there exists an abundance of open-sourced and legally available data on the internet that can be utilized for training. This implies the potential of employing MLLMs to annotate unstructured, unlabeled online data, thereby augmenting FL performance. Secondly, in contrast to the limited computational resources available on client devices, the server-side capabilities are significantly more robust. This disparity opens up the possibility of deploying additional, more powerful MLLM on the server side to provide assistance to the FL system. 
\begin{table*}[ht]
\vspace{-1cm}
\centering
\begin{tabular}{@{}lcccc@{}}
\toprule
Method & \begin{tabular}[c]{@{}c@{}}Multimodal\\ Supervision\end{tabular} & \begin{tabular}[c]{@{}c@{}}No Gradient\\ Upload for Privacy\end{tabular} & \begin{tabular}[c]{@{}c@{}}No Additional\\ Computing Burden \\ on Devices\end{tabular} & Compatibility \\
\midrule
CReFF \cite{shang2022federated},          &            &            & \checkmark &            \\
CLIP2FL \cite{shi2023clip}       & \checkmark &            &            &            \\
MLLM-LLaVA-FL (ours) & \checkmark & \checkmark & \checkmark & \checkmark \\
\bottomrule
\end{tabular}
\caption{Comparison between our methods and other status quo approaches for addressing long-tailed distribution challenges in FL}\label{table_diff}
\vspace{-0.3cm}
\end{table*}

Our framework has three key stages. The initial stage, termed Global Multimodal Pretraining, first employs MLLMs to generate descriptions for unlabelled data collected from the internet. Then we develop a novel pretraining strategy, Dynamic Weighted Pretraining (DWP), which enables MLLMs to assist the compact FL models within the FL framework to conduct pertaining more efficiently on the open-sourced dataset. In the second stage, known as Federated Finetuning, we distribute the pretrained FL model to clients for local training on their datasets similar to with traditional FL approaches. This stage is highly flexible and compatible, allowing the integration of various previously designed FL methods. During the third stage, we perform Global Alignment on the server-side aggregated FL model under MLLM supervision. This process, similar to the idea of alignment in large language models, is aimed at further refining the model's outputs to better align with task-specific requirements. Indeed, our framework is adaptable to a wide range of federated learning (FL) tasks. In this study, we specifically address the prevalent challenge of data heterogeneity in federated image classification tasks and the multimodal large language models we adopt here are the large vision-language models (LVLM). During the pretraining stage, thanks to the extensive open-vocabulary prior knowledge embedded in large vision-language models, these models are capable of generating detailed descriptions for complex images found on the internet. This pretraining process of our FL model with the assistance of LVLMs on a large dataset of text-images, enables our FL model to develop enhanced image representation capabilities to better counteract the effects of data heterogeneity inherent in FL environments. Furthermore, the global alignment stage can also be designed to mitigate the issue of long-tailed distributions, which tend to bias client-side models towards more frequently occurring classes. Our contribution can be summarized as follows.
\begin{itemize}
    \item Firstly, we pioneer the integration of the widely recognized multimodal large language model (MLLM) as an auxiliary tool in federated learning, aiming to enhance the utilization of previously underexplored internet data resources and server computational capabilities. Leveraging the formidable cross-modality representation capabilities and the vast open-vocabulary prior knowledge inherent in MLLMs, we introduce novel methodologies to address the challenges posed by long-tail distributions and data heterogeneity. 
    \vspace{-0.1cm}
    
    \item  In comparison to prevailing state-of-the-art approaches that address data heterogeneity in federated learning (FL), our methodology not only enhances privacy protection further but also significantly reduces the computational burden on client devices.
    \vspace{-0.1cm}
    \item Our extensive experimental results show that MLLM-LLaVA-FL can effectively handle heterogeneity and class-distribution imbalance, consistently surpassing the performance of existing state-of-the-art federated learning methodologies across a variety of datasets.
\end{itemize}

\section{Related Work}
\subsection{Multimodal large language model}
The introduction of GPT-4(Vision) \cite{openai-chatgpt}  and Gemini \cite{team2023gemini} have demonstrated remarkable abilities in Multimodal understanding and generation, sparking a research fervor on Multimodal large language model. This enthusiasm extends to a variety of tasks, including image-text comprehension \cite{li2023blip,liu2023llava,zhu2023minigpt}; video-text understanding \cite{li2023videochat,maaz2023video}; and audio-text understanding \cite{chu2023qwen}. Among them, recent studies in image-text comprehension with large vision-language models (LVLM) \cite{li2023blip,liu2023llava,zhu2023minigpt} have catalyzed notable advancements in harnessing the robust capabilities of large language models to tackle multimodal tasks effectively, such as crafting narratives from images and executing intricate reasoning tasks. Prominent instances include Visual ChatGPT \cite{wu2023visual}, which amalgamates diverse visual foundational models for intricate visual tasks and instructions, employing iterative feedback to synchronize visual and textual modalities. In a similar way, MM-REACT \cite{yang2023mm} merges ChatGPT with visual models for multimodal undertakings, especially in the Visual Question Answering (VQA) framework. BLIP-2 \cite{li2023blip}, notable for its Q-former model, has shown encouraging outcomes in VQA tasks, both in zero-shot and fine-tuning settings. LLaMA-Adapter \cite{gao2023llama} enhances multimodal fine-tuning efficiency by integrating adaptation prompt vectors as adjustable parameters, showcasing versatility in multimodal contexts. MiniGPT-4 \cite{zhu2023minigpt}, derived from GPT-4 and incorporating elements from BLIP-2 and Vicuna \cite{vicuna2023}, specializes in caption generation and model refinement through image-text pair fine-tuning. LLaVA \cite{liu2023llava}, leveraging GPT-4, focuses on a broad spectrum of instruction fine-tuning data, ranging from multi-turn QA to image descriptions, adopting a dual-stage fine-tuning approach that prioritizes language model loss while keeping the visual model static.

\subsection{Federated Learning with Heterogeneous Data}
Current methodologies tackling the challenge of data heterogeneity fall into the following broad categories. Some approaches aim to simultaneously improve the models on both clients and server sides through optimization techniques. Key contributions in this area have been made by the work of \cite{li2022effectiveness,huang2021behavior,huang2023rethinking,feng2023towards}, who have investigated various optimization methods. Other strategies focus on enhancing the stability of local models via knowledge transfer, a technique that is model-agnostic and has been explored in the research by \cite{xu2023learning,zhang2021parameterized,chen2020fedhealth}, aiming to mitigate data heterogeneity by spreading local training knowledge throughout the whole FL framework. Additional methods, such as those proposed by \cite{lin2020ensemble,chen2020fedbe}, concentrate on improving model aggregation on the server side to address data heterogeneity. Certain strategies also regulate the scheduling of client participation to avoid biasing the FL model towards classes that are more prevalent, as explored in the studies by \cite{yang2020federated,fedcbs}. While these approaches have advanced the handling of data heterogeneity, they often do not fully address the specific issues related to long-tailed distributions in FL.

A recent approach, CReFF \cite{shang2022federated}, introduces a decoupling strategy to create balanced class-distribution federated features for the server model and to retrain the classifier with these features. Nonetheless, CReFF encounters two main limitations due to its reliance on generating federated features through client-side gradient information: 1) The one-to-many relationship between gradients and samples can result in the problem becoming ill-posed; 2) The absence of semantic guidance might lead to federated features that lack discriminative ability for their respective classes. The subsequent attempt, CLIP2FL \cite{shi2023clip}, seeks to overcome these drawbacks by integrating a multimodal model to direct the federated learning process. However, it still has its own drawbacks. Firstly, deploying the sizable CLIP model on devices increases memory and computational demands. Secondly, transmitting both the gradient and parameters of local models to the server, as necessitated by both CLIP4FL and CReFF, raises significant privacy concerns, as attackers could potentially reconstruct client images through reverse engineering \cite{geiping2020inverting,zhu2019deep,haim2022reconstructing}.

\section{Methodology}
In the conventional federated learning (FL) pipeline, the FL models are typically assigned to local clients for training on their heterogeneous datasets. These models are then sent back to the server for aggregation. This cycle continues repeatedly until the FL training concludes. To better align the above FL framework with practical requirements, we incorporate the following two additional stages. The first stage occurs before local training, involving pretraining on the server side, a strategy supported by previous work \cite{chen2022pre,FedBERT} which found that pretraining can accelerate the convergence of FL training and mitigate the effects of data heterogeneity on convergence. The second stage takes place after aggregation, where the FL model may undergo further training to meet the broader requirements of FL companies, such as the performance and safety considerations we discuss later.

Drawing inspiration from recent works in learning paradigms of NLP~\cite{GPT4report,devlin2018bert,zhang2024sled,wang2024coreinfer,zhang2024min,kuo2023dacbert,joren2024sufficient,FedBERT,zhang2022join}, we structure our work into three parts: global multimodal pretraining, federated local finetuning, and global alignment. We deploy multimodal approaches at the server side to assist both global multimodal pretraining and global alignment phases. In this section, we will introduce our comprehensive framework as follows: Section 3.1 will delve into the details of our global multimodal pretraining; Section 3.2 will cover federated local finetuning; Section 3.3 will discuss global alignment; and in Section 3.4, we will compare our method with previous approaches.

\subsection{ Global Multimodal Pretraining}
\paragraph{Pretraining Dataset} 
As discussed in \cref{sec:intro}, the wealth of open-source multimodal data, such as images and their captions, remains underutilized resources for pretraining FL models. Often, these datasets are noisy, unlabeled or contain elements that are too complex, making them unsuitable for straightforward pretraining of the compact FL models. However, given the current advanced capabilities in multimodal processing of MLLMs, we now have new, convenient methods to leverage such data for pretraining purposes. Utilizing GPT-4, akin to the approach used in LLaVA, we can transform complex image data collected from the internet into three main categories:
\begin{itemize}
    \item \textit{Conversation}: This category includes dialogues between an assistant and an individual seeking specific information about a photo. The assistant's responses simulate observing the image directly, answering a variety of questions about the visual content, such as identifying objects, counting them, describing actions, pinpointing locations, and noting their spatial relations.
    
    \item \textit{Detailed Description}: To gain a thorough understanding of an image, we formulated a series of questions designed to elicit detailed descriptions. Responses to these questions were generated using GPT-4, enriching our dataset with nuanced insights into the images.
    
    \item \textit{Complex Reasoning}: This category focuses on more sophisticated reasoning questions based on the content of the images. Answering these questions involves a detailed logical breakdown, reflecting a deep comprehension of the images and the ability to reason through them.
\end{itemize}

Leveraging the aforementioned dataset formulations, we are equipped to facilitate the pretraining of FL models with the support of Multimodal Large Language Models.

\paragraph{Pretraining Mechanism}
\begin{figure}[t]
    \centering
    \includegraphics[width=0.9\linewidth]{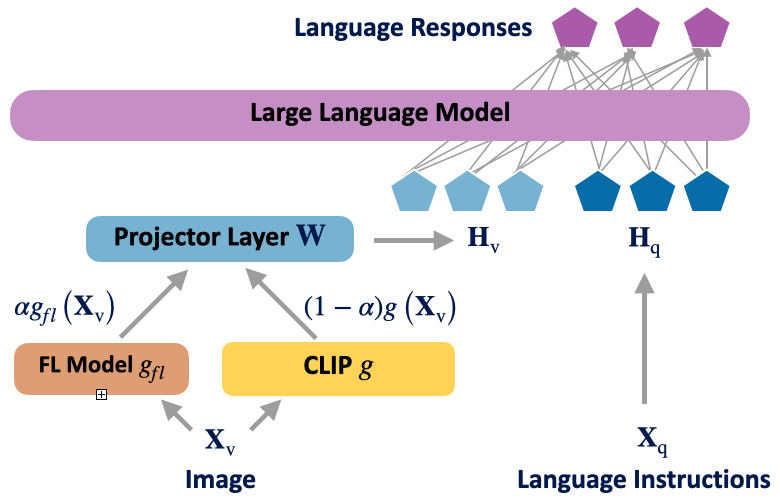}
    \caption{The visualization of our pretraining mechanism}
    \label{fig:framework}
\end{figure}

Our pretraining mechanism draws inspiration from the structure of LLaVA, a highly effective and recent multimodal large language model. It consists of three key components: a visual encoder $g$, which is a frozen pretrained CLIP model; a projection layer designed to align the features of the visual model with the text domain embeddings, where the projection layer is a trainable matrix $W$; and a part comprising a large language model (LLM), typically employing promising models such as Vicuna or LLaMA-2. The workflow of LLaVA proceeds as follows: For the data formats mentioned earlier, whether it be a conversation, detailed description, or complex reasoning, the input includes a text modality instruction $\mathbf{X}_{\mathrm{q}}$ (e.g., "Could you provide a detailed description of this image?") and an image $\mathbf{X}_{\mathrm{v}}$. The instruction $\mathbf{X}_{\mathrm{q}}$ passes through an embedding layer to obtain the text embedding $\mathbf{H}_q$. As for the image $\mathbf{X}_{\mathrm{v}}$, it first goes through the visual encoder $g$ to acquire the grid feature $\mathbf{Z}_{\mathrm{v}}$, which then passes through the projection layer to obtain the visual embedding $\mathbf{H}_v$, aligned in dimension with $\mathbf{H}_q$:
\begin{equation}
    \mathbf{H}_v=\mathbf{W} \cdot \mathbf{Z}_{\mathrm{v}} \text{, where } \mathbf{Z}_{\mathrm{v}}=g\left(\mathbf{X}_{\mathrm{v}}\right) 
\end{equation}
The text embedding $\mathbf{H}_q$ and the visual embedding $\mathbf{H}_v$ are concatenated and subsequently input into the LLM. The resulting output is the MLLM's response to the given inputs. Throughout the LLaVA training process, both the projector and the LLM are trainable, whereas the visual encoder and the CLIP model remain fixed and are not subject to training.

Drawing from the LLaVA mechanism, our Global Multimodal Pretraining essentially integrates the compact FL model, designed for downstream image classification tasks, as a component within the LLaVA framework's visual encoder. The FL model is made trainable and is denoted as $g_{fl}$. Inspired by the previous work in knowledge distillation\cite{hinton2015distilling,mirzadeh2020improved,jin2019knowledge,zhang2023reaugkd} and Bayesian Algorightms \cite{zhang2020variance,chen2022we,welling2011bayesian,zhang2019cyclical,zhang2020stochastic,liu2019stein,zhao2018selfadversarially}, We have developed an approach termed Dynamic Weighted Distillation, which involves computing a weighted average of the visual features obtained from the FL model and those from the original visual encoder:
\begin{equation}
\mathbf{Z}_{\mathrm{v}}= (1-\alpha) g\left(\mathbf{X}_{\mathrm{v}}\right) + \alpha g_{fl}\left(\mathbf{X}_{\mathrm{v}}\right)
\end{equation}

In this equation, $\mathbf{Z}_{\mathrm{v}}$ represents the weighted combined visual features, $g\left(\mathbf{X}_{\mathrm{v}}\right)$ indicates the visual features from the original visual encoder, $g_{fl}\left(\mathbf{X}_{\mathrm{v}}\right)$ refers to the visual features from the FL model, and $\alpha$ is the dynamic weighting factor that adjusts the influence of each feature. During the pretraining phase, the CLIP model $g$, the projector $W$, and the LLM are kept static, with only $g_{fl}$, the FL model component, being trainable. Initially, $\alpha$ is set to 0, and as pretraining progresses, it gradually increases to 1, where it remains for the duration of the pretraining. This approach is termed Dynamic Weighted Pretraining. The rationale behind this strategy stems from the typically smaller size of the FL model compared to the original CLIP visual encoder. This size discrepancy is due to the constraints imposed by subsequent local training on edge devices within the FL framework. Directly substituting the large CLIP model with a more compact FL model could significantly hamper the process of multimodal alignment, owing to the vast difference in capacity between the two visual models resulting from their size difference.

\subsection{Federated Finetuning}
In this subsection, we delve into the Federated Finetuning phase. Upon obtaining a pretrained FL model $g_{fl}$ from the initial stage, we append classifier layers to it, tailoring the model for image classification tasks on local datasets on the client side. If we denote the set of all model parameters for the $k$-th client at the $t$-th local step as $w_k^t$, and the local data as $\mathcal{D}^k$, then the $k$-th client updates the received model in a manner similar to FedAvg:
\begin{equation}
w_k^{t+1} \leftarrow w_k^t - \eta \nabla_w L_{\text{loc}}(w^t; \mathcal{D}^k)
\end{equation}, where the $L_{\text{loc}}$ represents the local loss function.

After local training, the model parameters $w_k$ are sent back to the server for global aggregation, where we also utilize FedAvg:
\begin{equation}
    w_{agg}^{t+1}=\sum_{k \in \Omega^t} \frac{\left|\mathcal{D}^k\right|}{\sum_{k \in \Omega^t}\left|\mathcal{D}^k\right|} w_k^{t+1}
\end{equation}, where the $\Omega^t$ is the set of clients selected at the $t$-th round.
 
This stage mirrors the traditional FL framework closely. Drawing inspiration from learning paradigms in NLP, we refer to this phase as Federated Finetuning, in light of the pretraining conducted in the preceding step. It's important to note the flexibility and compatibility of our framework; we can substitute FedAvg with any other existing FL methods designed to enhance local training and global aggregation, aiming to boost final utility performance metrics like accuracy, or system performance aspects like speed or computational efficiency. Furthermore, this phase does not cause additional privacy concerns, and existing methods for privacy or safety protection can be seamlessly integrated.

\subsection{Global Alignment}
Recent studies on the alignment of large language models, such as Reinforcement Learning from Human Feedback (RLHF), are focused on refining the model's outputs to more closely resonate with human-like understanding and reasoning. This enhancement significantly improves the model's capability in tasks that require the interpretation and execution of complex instructions. In a similar vein, companies engaged in federated learning (FL) have analogous requirements for models after global aggregation. For instance, concerning safety requirements, an FL company must ensure that models trained via federated learning do not leak user information. Additionally, there are performance-related requirements, such as adjusting the model to prevent biases caused by long-tailed distributions or training the model on new datasets to acquire new skills. Typically, this involves constructing an alignment dataset $\mathcal{D}_{align}$ and selecting a suitable global alignment function $L_{align}$:
\begin{equation}
w_{agg}^{new} \leftarrow w_{agg} - \eta \nabla_w L_{align}(w_{agg}; \mathcal{D}_{align})
\end{equation}

To address the issue of long-tailed distributions, one could design $\mathcal{D}_{align}$ as a small, class-balanced dataset encompassing all categories, with $L_{align}$ defined as follows:
\begin{equation}
L_{align}=L_{ce}\left(y, p\right)+\beta \cdot K L\left(q \| p\right),
\end{equation}
where $L_{ce}(\cdot, \cdot)$ is the cross-entropy loss. $y$ is the label and $p$ is the output logits vector of the FL models. Since the pretrained CLIP model in the pertaining mechanism has zero-shot image classification capability \cite{radford2021learning}, $q$ denotes the output logits vector of the CLIP model. $K L$ is the Kullback-Leibler divergence and $\beta$ is a hyperparameter balanced these two losses.

\begin{table*}[ht]
\centering
\vspace{-0.5cm}
\resizebox{0.8\textwidth}{!}{%
\renewcommand{\arraystretch}{1.2}
\begin{tabular}{llcccccc}
\hline 
\multirow{2}{*}{Type} & \multirow{2}{*}{Method} & \multicolumn{3}{c}{CIFAR-10-LT} & \multicolumn{3}{c}{CIFAR-100-LT} \\
\cline{3-8} 
& & IF=100 & IF=50 & IF=10 & IF=100 & IF=50 & IF=10 \\
\hline 
\multirow{7}{*}{Heterogeneity-oriented FL methods} & FedAvg & 56.17 & 59.36 & 77.45 & 30.34 & 36.35 & 45.87 \\
& FedAvgM & 52.03 & 57.11 & 70.81 & 30.80 & 35.33 & 44.66 \\
& FedProx & 56.92 & 60.89 & 76.53 & 31.67 & 36.30 & 46.10 \\
& FedDF & 55.15 & 58.74 & 76.51 & 31.43 & 36.22 & 46.19 \\
& FedBE & 55.79 & 59.55 & 77.78 & 31.97 & 36.39 & 46.25 \\
& CCVR & 69.53 & 71.89 & 78.48 & 33.43 & 36.98 & 46.88 \\
& FedNova & 57.79 & 63.91 & 77.79 & 32.64 & 36.62 & 46.75 \\
\hline 
\multirow{3}{*}{Imbalance-oriented FL methods} & Fed-Focal Loss & 53.83 & 57.42 & 73.74 & 30.67 & 35.25 & 45.52 \\
& Ratio Loss & 59.75 & 64.77 & 78.14 & 32.95 & 36.88 & 46.79 \\
& FedAvg+ $\tau$-norm & 49.95 & 51.41 & 72.08 & 26.22 & 33.71 & 43.65 \\
\hline 
Classifier-retraining & CReFF & 70.55 & 73.08 & 80.71 & 34.67 & 37.64 & 47.08 \\
SOTA & CLIP2FL & 73.37 & 75.35 & 81.18 & 37.56 & 41.29 & 48.20  \\
\hline 
\multirow{2}{*}{Our framework} & \multirow{2}{*}{MLLM-LLaVA-FL} & \textbf{75.49} & \textbf{76.11} & \textbf{81.45} & \textbf{39.50} & \textbf{42.34} & \textbf{48.87} \\
& & $(\textcolor{green}{\uparrow 2.12})$ & $(\textcolor{green}{\uparrow 1.24})$ & $(\textcolor{green}{\uparrow 0.27})$ & $(\textcolor{green}{\uparrow 1.94})$ & $(\textcolor{green}{\uparrow 1.05})$ & $(\textcolor{green}{\uparrow 0.67})$ \\
\hline
\end{tabular}%
}
\caption{Top-1 classification accuracy(\%) on CIFAR-10-LT and CIFAR-100-LT datasets with different FL methods, where th results are referred in \cite{shang2022federated,shi2023clip}. The best results are marked in bold.}\label{cifar-results} 
\vspace{-0.5cm}
\end{table*} 
The idea of using a class-balanced dataset to alleviate the challenges of long-tailed distributions aligns with the concepts of data resampling in centralized training to handle class imbalance and client selection in federated learning. The feasibility of such an alignment dataset $\mathcal{D}_{align}$ existing on the server side is justifiable in most cases because, in practice, for global aggregation, the server typically predefines the categories for model classification and organizes them, which is essential for subsequent federated learning processes. Otherwise, the global aggregation of classifier layers would become chaotic. Knowing the categories, companies could feasibly collect data from the internet or generate data using powerful image-generation models like Stable Diffusion or Midjourney. However, we acknowledge that in some extreme cases, data collection can be challenging, necessitating the design of more specific $L_{align}$ and $\mathcal{D}_{align}$, which we leave for future work.

\section{Experiment}
\subsection{Experiment Setup}

\paragraph{Dataset$\&$Implementation}
We applied our MLLM-LLaVA-FL framework to three widely-used long-tailed datasets: CIFAR-10/100LT \cite{cifar10} and ImageNet-LT \cite{liu2019large}. As for the first two datasets, we adopt the same sampling technique as previous studies \cite{cui2019class} to create long-tailed distributions with various imbalance factors $(\mathrm{IF}=100,50,10)$, and we follow CReFF \cite{shang2022federated} to use Dirichlet distribution with the key parameter $\alpha$ to generate the heterogeneous data partition among clients, where the value of $\alpha$ is set to 0.5 on CIFAR-10/100-LT. ImageNet-LT has $115.8 \mathrm{~K}$ images from 1000 classes and the number of images per class ranging from 1280 to 5, where the value of $\alpha$ is set to 0.1. We utilized ResNet-8 as the feature extractor for CIFAR-10/100-LT and ResNet-50 for ImageNet-LT, adding an MLP layer to each to align their feature dimensions with CLIP's outputs. The number of clients is set to 20, and we select 40$\%$ at random for each training round. The client-side training batch size was uniform at 32 across Cifar-10/100 and imagenet. All the above settings are the same as the previous work in \cite{shi2023clip}. We employed the standard cross-entropy loss by default and executed 200 communication rounds. For the pertaining part, we adopt the pertaining dataset of LLaVA, CC-595K, and train the model for 4 epochs with a learning rate of 2e-3 and a batch size of 128. During the first 2 epochs, the $alpha$ in our pretraining mechanism increase from 0 to 1 following the cosine scheduler and then the value remains 1 for the following epochs. All the experiments were conducted using PyTorch on a single Nvidia A100 80G GPU. 
\paragraph{Baselines} We compare MLLM4FL with 13 FL methods: FedAvg \cite{mcmahan2017communication}, FedAvgM \cite{hsu2019measuring}, FedProx \cite{li2018federated}, FedDF \cite{FedED-2020}, FedBE \cite{chen2020fedbe}, CCVR \cite{luo2021no} and FedNova \cite{wang2020tackling}, Fed-Focal Loss \cite{sarkar2020fed}, Ratio Loss \cite{wang2020addressing} and FedAvg with $\tau$-norm \cite{kang2019decoupling}, CReFF \cite{shang2022federated} and CLIP2FL \cite{shi2023clip}. The first seven approaches are heterogeneity-oriented , and Fed-Focal Loss, Ratio Loss and FedAvg with $\tau$-norm are imbalance-oriented.

\subsection{Experimental Results}
Results for CIFAR-10/100-LT are presented in \cref{cifar-results}, where we evaluate the performance of our CLIP2FL against a range of FL approaches on both CIFAR-10-LT and CIFAR-100-LT datasets. Notably, MLLM-LLaVA-FL outperforms other methods in terms of classification accuracy on both datasets. Specifically, at an Imbalance Factor (IF) of 100, which presents a severe imbalance,  MLLM shows an improvement of 2.12\% and 1.94\% in classification accuracy over CLIP2FL for CIFAR-10-LT and CIFAR-100-LT, respectively. Under the condition of IF = 50 or 10, MLLM still manages to enhance performance by around 1\%. This underscores MLLM-LLaVA-FL's effectiveness and its outperforms over competing methods to deal with heterogenous and long-tailed distributions.

\begin{table*}[ht]
\centering
\resizebox{0.8\textwidth}{!}{%
\renewcommand{\arraystretch}{1.2}
\begin{tabular}{llcccc}
\hline 
\multirow{2}{*}{Type} & \multirow{2}{*}{Method} & \multicolumn{4}{c}{ImageNet-LT} \\
\cline{3-6} 
& & All & Many & Medium & Few \\
\hline 
\multirow{5}{*}{Heterogeneity-oriented FL methods} & FedAvg & 23.85 & 34.92 & 19.18 & 7.10 \\
& FedAvgM & 22.57 & 33.93 & 18.55 & 6.73 \\
& FedProx & 22.99 & 34.25 & 17.06 & 6.37 \\
& FedDF & 21.63 & 31.78 & 15.52 & 4.48 \\
& CCVR & 25.49 & 36.72 & 20.24 & 9.26 \\
\hline 
\multirow{3}{*}{Imbalance-oriented FL methods} & Fed-Focal Loss & 21.60 & 31.74 & 15.77 & 5.52 \\
& Ratio Loss & 24.31 & 36.33 & 18.14 & 7.41 \\
& FedAvg+ $\tau$-norm & 21.58 & 31.66 & 15.76 & 4.33 \\
\hline 
Classfier-retraining & CReFF & 26.31 & \textbf{37.44} & 21.87 & 10.29 \\
\hline 
\multirow{2}{*}{Our framework} & \multirow{2}{*}{MLLM-LLaVA-FL} & \textbf{27.53} & {30.85} & \textbf{25.89} & \textbf{25.58} \\
& & $(\textcolor{green}{\uparrow 1.22})$ & $(\textcolor{red}{\downarrow 6.59})$ & $(\textcolor{green}{\uparrow 4.02})$ & $(\textcolor{green}{\uparrow 15.29)}$ \\
\hline
\end{tabular}%
}
\caption{Top-1 accuracy(\%) on ImageNet-LT dataset with different FL method}\label{imagenet-results}
\vspace{-0.1cm}
\end{table*}

In the context of ImageNet-LT, \cref{imagenet-results} presents a comparison of the accuracy achieved by our MLLM-LLaVA-FL framework against various FL approaches. The evaluation is segmented into four groups based on the number of samples per class: “Many” (over 100 samples), “Medium” (20 to 100 samples), “Few” (less than 20 samples), and “All” (overall accuracy). While our method may not fully match the performance of CReFF in the “Many” categories, it excels in “Overall” accuracy and the “Medium” and "Few" category. These results underscore MLLM-LLaVA-FL's capability not just in enhancing overall model performance but also in significantly improving classification outcomes for categories with fewer samples. The ImageNet-LT results highlight the effectiveness of MLLM-LLaVA-FL in tackling the inherent challenges of long-tailed data distributions.

\subsection{Further Analysis}
We conduct some further analysis to verify the effectiveness of our framework, especially the importance of global pertaining and global alignment.
\begin{table*}[]
    \centering
    \resizebox{0.8\linewidth}{!}{
    \begin{tabular}{
    >{\centering\arraybackslash}p{2cm}
    >{\centering\arraybackslash}p{3cm}
    >{\centering\arraybackslash}p{2cm}
    >{\centering\arraybackslash}p{2cm}
    >{\centering\arraybackslash}p{2cm}}
    \hline
    \textbf{Dataset} & \textbf{Initialization} & \textbf{0.4\%} & \textbf{1\%} & \textbf{2\%} \\
    \hline
    Cifar10 & w/ Pretraining & \textbf{(3, 37.62)} & \textbf{(5, 39.46)} & \textbf{(3, 37.83)} \\
    & w/o Pretraining & (6, 30.44) & (10, 29.8) & (14, 31.99) \\
    Cifar100 & w/ Pretraining & \textbf{(5, 23.91)} & \textbf{(6, 24.46)} & \textbf{(6, 24.28)} \\
    & w/o Pretraining & (10, 18.61) & (15, 20.27) & (11, 19.42) \\
    \hline
    \end{tabular}
    }
\caption{Comparison between pretrained and non-pretrained models under constrained training dataset settings. The format is (number of epochs, highest accuracy)}\label{table_fewshot}
\end{table*}
\subsubsection{Ablation studies on our pretraining mechanism} To evaluate the effectiveness of pretraining, we conducted a comparative analysis between a pretrained model and a non-pretrained model under a constrained training dataset scenario akin to few-shot learning, aiming to mirror real-world conditions where the amount of data available on each device is limited. We generated subsets of the CIFAR-10 and CIFAR-100 datasets at varying proportions and trained both models for 30 epochs. Our findings, detailed in \cref{table_fewshot}, outline the number of epochs required by each model to reach predetermined accuracy thresholds with different training sample sizes, along with the highest accuracy achieved by each model within the 30-epoch span.

\begin{figure}[ht]
    \centering
    \includegraphics[width=0.9\linewidth]{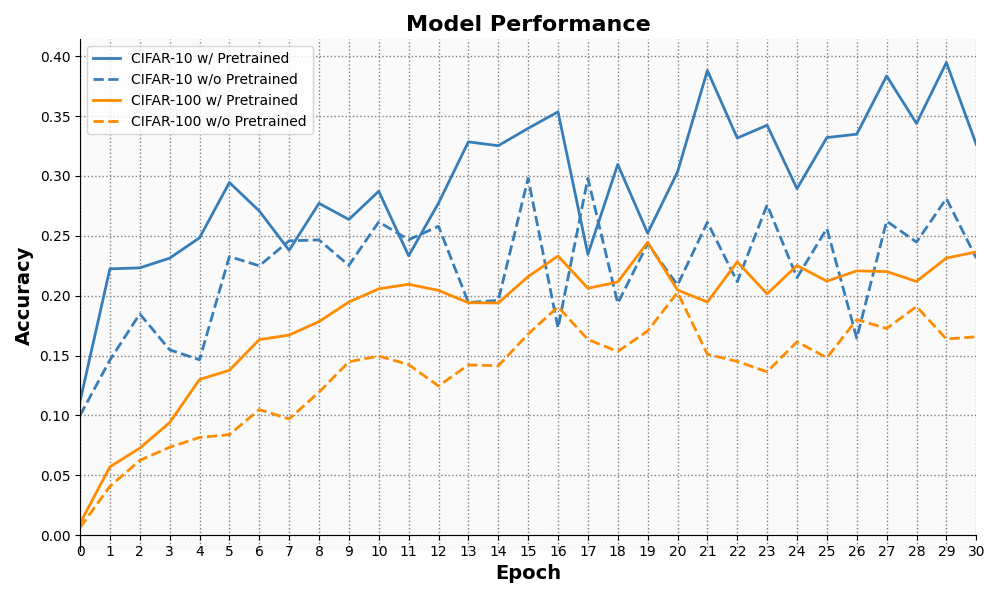} 
    \caption{The comparative analysis of pretrained and non-pretrained models using 1\% subsets of CIFAR-10/100 training data.}
    \label{fig:fewshot_acc}
    \vspace{-0.5cm}
\end{figure}

Specifically, within the CIFAR-10 context, to attain a target accuracy of $25\%$, the pretrained model consistently outperformed the non-pretrained model, requiring fewer epochs across all sample sizes. Similarly, for the CIFAR-100 dataset, in pursuit of a target accuracy of $15\%$, the pretrained model proved more efficient, also necessitating fewer epochs. \cref{fig:fewshot_acc} further illustrates the superiority of our pretraining approach, demonstrating that models pretrained using our methodology surpass those trained from scratch in terms of accuracy. 

\begin{figure*}[ht]
    \centering
    \includegraphics[width=0.7\textwidth]{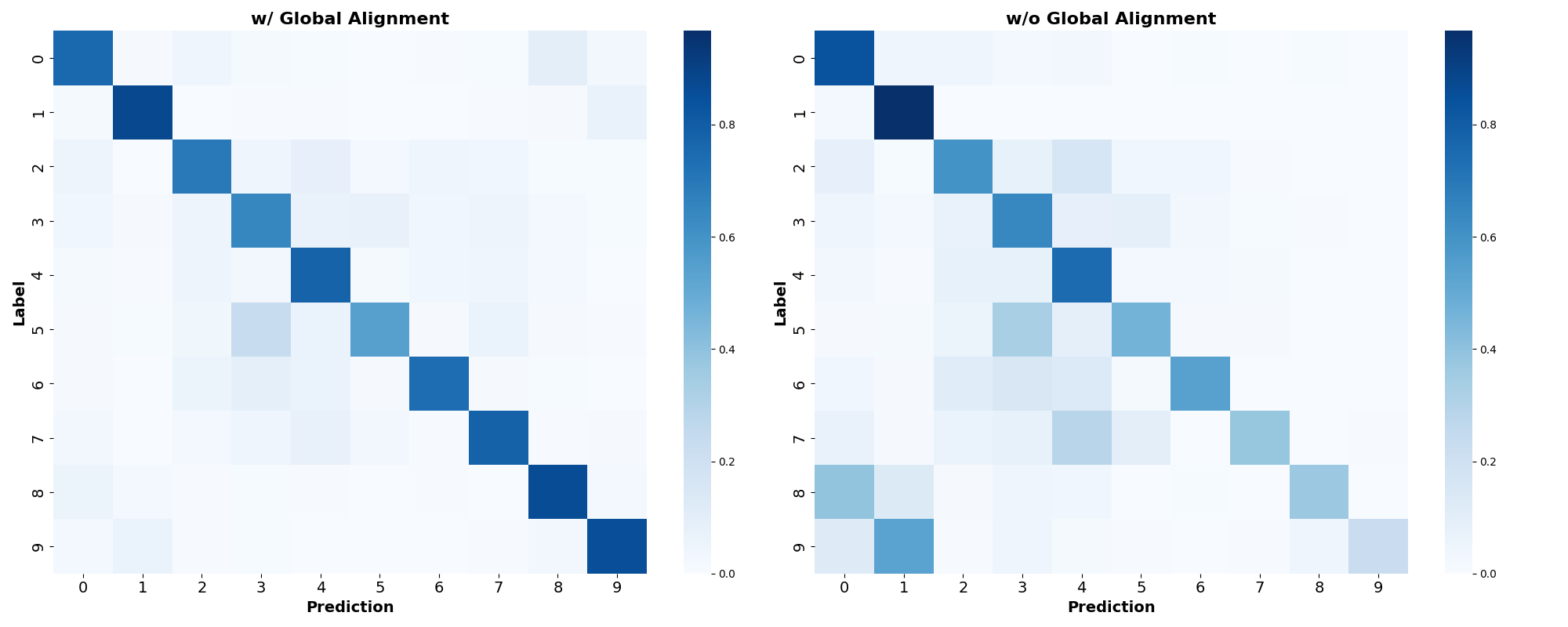}
    \caption{The comparative analysis of aligned and non-aligned models with normalized confusion matrices.}
    \label{fig:confusion}
    \vspace{-0.4cm}
\end{figure*}
\subsubsection{Ablation studies on our global alignment mechanism} In \cref{fig:confusion}, we present the confusion matrix for our model equipped with global alignment within the CIFAR-10-LT dataset, characterized by an imbalance factor of 100. We normalize this confusion matrix by the volume of data in each class. Additionally, we illustrate a normalized confusion matrix for a baseline model devoid of global alignment within our Federated Learning (FL) framework. The visual representations indicate that, with alignment in place, data from each class can be accurately classified. In contrast, the absence of alignment yields inferior results, particularly for classes with few data. Compared to classes with abundant data, those with fewer instances often experience misclassification, with minority class data being inaccurately labeled as belonging to majority classes. This starkly underscores the significance of our global alignment strategy in enhancing both the performance and fairness of the FL system.
 \vspace{-0.3cm}
\subsubsection{Discussion}
 \paragraph{Privacy:}Our strategy sidesteps the conventional requirement for clients to send gradients back to the server, as seen in methods like CReFF \cite{shang2022federated} and CLIP2FL \cite{shi2023clip}. This aspect is vital because the transmission of gradients could enable the server to perform reverse engineering attacks \cite{geiping2020inverting,zhu2019deep,haim2022reconstructing,tang2022fade}, potentially endangering client data confidentiality. By eliminating this step, our method diminishes the likelihood of leaking sensitive client information, promoting a more secure and privacy-centric learning environment.
 \vspace{-0.3cm}
\paragraph{Computational Efficiency:}Our approach also stands out for its computational economy. Contrary to approaches like CLIP2FL, which necessitate deploying sizable multimodal models such as CLIP on client devices—demanding significant memory and potentially being unfeasible for edge devices with limited resources—our method positions the MLLM solely on the server side. At the client level, we deploy only the compact FL models. This resolution not only addresses memory constraints but also reduces the time and energy expenditure associated with federated local training. 
Consequently, our framework is rendered more practical and appealing for an extensive array of devices, particularly those with restricted storage capacities.
 \vspace{-0.3cm}
\paragraph{Compatibility:}Our approach stands out for its adaptability, unlike specific methodologies like CReFF and CLIP2FL that impose unique requirements on federated local training and global aggregation. Our framework can be compatible with a wide array of existing FL algorithms. This includes but is not limited to, client selection strategies and various techniques aimed at further enhancing client privacy protection. For a more comprehensive comparison, please refer to the \cref{table_diff} highlighting the advantages of our method.
\section{Conclusion}
To overcome the challenges of federated learning in the context of heterogeneous and long-tailed data distributions, we introduced a novel framework, MLLM-LLaVA-FL. This framework is structured around three core stages: global pretraining, federated fine-tuning, and global alignment. This marks the inaugural integration of MLLMs into an FL system. Leveraging the strong multimodal capacities of MLLMs, our approach taps into the vast yet previously underutilized reservoir of open-source data available online, alongside substantial server-side computational resources. Crucially, our methodology does not compromise privacy nor impose additional computational demands on client devices. Experimental evidence verifies the efficacy of our framework, paving the way for future research on more multimodal tasks.
\vspace{-0.3cm}
\paragraph{Acknowledgement} This research was supported in part by the memberships of the NSF IUCRC 1822085, the Center for Alternative Sustainable and Intelligent Computing (ASIC). 
{\small
\bibliographystyle{ieee_fullname}
\bibliography{egbib}
}

\end{document}